\documentclass[letterpaper]{article}
\usepackage{aaai}
\usepackage{times}
\usepackage{helvet}
\usepackage{courier}
\usepackage{hyperref}
\usepackage[ruled,longend,vlined,linesnumbered]{algorithm2e}

\begin{document}
%
\title{Totally and Partially Ordered Hierarchical Planners in PDDL4J Library}
\author{Damien Pellier, Humbert Fiorino\\
Univ. Grenoble Alpes - LIG\\
F-38000 Grenoble, France \\
\{Damien.Pellier, Humbert.Fiorino\}@imag.fr
}
\maketitle
\begin{abstract}
\begin{quote}
In this paper, we outline the implementation of the TFD (Totally Ordered Fast Downward) and the PFD (Partially ordered Fast Downward) hierarchical planners that participated in the first HTN IPC competition in 2020. These two planners are based on forward-chaining task decomposition coupled with a compact grounding of actions, methods, tasks and HTN problems.
\end{quote}
\end{abstract}

\section{Introduction}

The TFD (Totally Ordered Fast Downward) and PFD (Partial ordered Fast Downward) hierarchical planners are based on forward-chaining task decomposition used by the SHOP2 planner \cite{nau03} coupled with a compact grounding of actions, methods, tasks and HTN problems. Both planners accept as input HDDL (Hierarchical Domain Description Language) proposed by \cite{holler20} and are implemented on top of the PDDL4J library \cite{pellier18}. In this short paper we present first the compact representation used by TFD and PFD as well as the grounding procedure implemented. Finally, we conclude with a brief presentation of the search strategy implemented in both planners.

\section{Grounding technique}

Most modern planners work with grounded representations of the planning problem. However, planning domains are commonly defined with a lifted description language such as PDDL \cite{mcdermott98} or HDDL \cite{holler20}. Thus, planning systems have to generate a grounded representation of the lifted domain in a preprocessing step, the objective of which is to generate the most compact grounded representation possible without removing any action, method or fluent needed for a solution plan. The more compact the grounded representation is, the more efficient is the search for a solution plan as reducing the size of the search space speeds up search and heuristic value computation. In practice, computing a grounded representation from a lifted representation is quite straightforward. All possible instantiations of ground predicates, primitive actions, abstract tasks and methods must be computed, and appropriately replaced by their ground versions in the lifted representation.

In the context of classical (non hierarchical) planning, the planners FF \cite{hoffmann01} and FastDownward \cite{helmert06} have implemented techniques for transforming lifted to ground planning representation that are still used in many planners today. Regarding hierarchical planning, \cite{ramoul17} inspired from \cite{koehler99} have been the first to propose an efficient grounding preprocessing, and it was successfully applied to the planners proposed by \cite{schreiber19}. Recently, \cite{Behnke20} has proposed novel techniques.

In the TFD and PFD planners, the grounding combines the approches proposed by \cite{ramoul17} and \cite{Behnke20}. It is based on 6 steps:
\begin{enumerate}
    \item Encode the lifted domain into an integer representation,
    \item Simplify the lifted representation and infer types from predicates,
    \item Instantiate the set of actions by removing unreachable actions with respect to the inertia principle \cite{koehler99},
    \item Filter out actions grounded in reachability analysis \cite{hoffmann01},
    \item Instantiate the set of methods by removing unuseful methods with respect to the inertia principle, and by recursively decomposing the initial tasks network of the planning problem. Tasks that are not reachable, i.e., tasks for which there are no relevant ground actions or methods are pruned.
    \item Encode the actions and the methods into bitset representation.
\end{enumerate}

More details on the implementation of the different instantiation techniques are available in PDDL4J opensource repository: \url{https://github.com/pellierd/pddl4j}.

\section{Search procedures}

The non-deterministic TFD procedure for solving a HTN planning problem is given in Algorithm~\ref{TFD}. This procedure is based directly on the recursive definition of a solution plan for HTN planning problems.

The TFD procedure takes as input a problem $P = (s_0,T,A,M)$ where $s_0$ is the initial state, $T = \langle t_1,t_2,...,t_k\rangle$ is a list of tasks, $A$ the set of actions, and $M$ the set of methods, all in their ground representation. First, the procedure tests if the list of tasks $T$ is empty (line~2). In this case, no task has to be executed, thus the empty plan is returned. Then the procedure gets the first task $t_1$ of the list $T$. Two cases must be considered depending on the type of $t_1$:
\begin{description}
\item[Case 1.] If $t_1$ is primitive (line~3) then the procedure computes the set of all the ground actions that accomplishes $t_1$ and that are applicable in $s_0$ (line 4). If there is no action (line~5), the procedure fails because no action accomplishes the goal task $t_1$. Then the procedure non-deterministically chooses an action that accomplishes the task (line~6), and calls itself recursively on the planning problem $P' =  (\gamma(s_0, a_1), T - \{t_1\}, A, M)$ (line~7). Finally, if the recursive call to the procedure fails to find a plan $\pi$, it returns failure (line~8); otherwise it returns the plan that is the concatenation of $a$ and $\pi$ (line~9).
\item[Case 2.] If $t_1$ is non-primitive (line 10) then the procedure computes the set of ground decompositions that accomplish $t_1$ and that are applicable in $s_0$ (line~11). If there is no decomposition to accomplish $t_1$ (line~12) then the procedure returns failure. Otherwise the procedure non-deterministically chooses a decomposition $d$ that accomplishes the task $t_1$ (line~13), and recursively returns the solution plan for the problem
$P' = (s_0, \mbox{\it subtasks}(d) \oplus \langle t_2, \ldots, t_k \rangle, A, M)$ (line~14).
\end{description}

Practically non-deterministic choices are made by systematically choosing the task networks with the least amount of non-decomposed tasks. In the case where several networks have the same number of tasks remaining to be decomposed, the task network containing the least number of actions is chosen.

The search procedure implemented in PFD is almost similar. The main difference is no longer to choose the first $t_1$ task in the task network but to choose the first task that does not have any predecessor task in the task network. In addition, each time case 2 applies, it is necessary to check the consistency of the ordering constraints of the task network in order to generate a-cyclic task networks. This check is performed before line 14 by calculating the transitive closure of the ordering constraints. The computation of the transitive closure is based on Warshall algorithm. The complexity is $O(n)$ where $n$ is the number of tasks of the task network.

\begin{algorithm}[t]
\DontPrintSemicolon
\label{TFD}
\caption{TFD($s_0,T,A,M$)}
Let $T= \langle t_1, \ldots, t_k\rangle$ \;
\lIf{$k = 0$}{return the empty plan $\langle \rangle$}
\If{$t_1$ is primitive}{
	$A' \leftarrow$ the set of relevant actions for $t_1$  and applicable in $s_0$\;
    \lIf{$A' = \emptyset$}{\Return {\bf failure}}
    non-deterministically choose an action $a \in R
    A'$\;
    $\pi \leftarrow TFD(\gamma(s_0, a), \langle t_2, \ldots, t_k \rangle, A, M)$\;
    \lIf{$\pi =$ {\bf failure}}{return {\bf failure}}
	\lElse{\Return $a \oplus \pi$}
} \uElseIf{$t_1$ is a non-primitive task}{

	$M' \leftarrow$ the set of relvant methods for $t_1$ and applicable in $s_0$ \;

    \lIf{$M' = \emptyset$}{\Return {\bf failure}}

    non-deterministically choose a decomposition $m \in M'$\;
    \Return~TFD($s_0,$ {\it subtasks}$(d) \oplus~\langle t_2, \ldots, t_k \rangle, A, M$)
}
\end{algorithm}

\bibliographystyle{aaai}
\bibliography{ref}

\end{document}